%% file: GlobComm.tex
\documentclass[conference]{IEEEtran}

\usepackage{amsmath,amsthm,amssymb}
\usepackage{mathrsfs,mathtools,mathabx}
\usepackage{bm}
\usepackage{graphicx}
\usepackage{textcomp}
\usepackage{xcolor}
\usepackage{dsfont}
\usepackage{algorithm,algpseudocode}
\usepackage{algpseudocode}
\usepackage{pgfplots}
\pgfplotsset{compat=newest}
\usepackage{subfigure}
\usepackage{float}
\usepackage{xparse}
\usepackage[bookmarks=false, hidelinks]{hyperref}
\usepackage{cite}
\usepackage{tikz}
\usetikzlibrary{arrows,decorations.pathreplacing,patterns}

\newtheorem{rem}{Remark}

\theoremstyle{definition}

\newcommand{\TODO}[1]{{\color{black} #1}}

\allowdisplaybreaks 
\usepackage{balance} 

\definecolor{chestnut}{rgb}{0.8, 0.36, 0.36}
\definecolor{airforceblue}{rgb}{0.36, 0.54, 0.66}
\definecolor{cadmiumorange}{rgb}{0.93, 0.53, 0.18}
\definecolor{bleudefrance}{rgb}{0.19, 0.55, 0.91}
\definecolor{carolinablue}{rgb}{0.6, 0.73, 0.89}
\definecolor{blue(ncs)}{rgb}{0.0, 0.53, 0.74}
\definecolor{dodgerblue}{rgb}{0.12, 0.56, 1.0}
\definecolor{cssgreen}{rgb}{0.0, 0.5, 0.0}
\definecolor{cadmiumgreen}{rgb}{0.0, 0.42, 0.24}
\definecolor{cadmiumorange}{rgb}{0.93, 0.53, 0.18}
\definecolor{amaranth}{rgb}{0.9, 0.17, 0.31}
\definecolor{bluegray}{rgb}{0.4, 0.6, 0.8}
\definecolor{cadmiumgreen}{rgb}{0.0, 0.42, 0.24}
\definecolor{amaranth}{rgb}{0.9, 0.17, 0.31}
\definecolor{amethyst}{rgb}{0.6, 0.4, 0.8}
\definecolor{amber}{rgb}{1.0, 0.75, 0.0}

\definecolor{babyblue}{rgb}{0.54, 0.81, 0.94}

\begin{document}

\title{Blind Asynchronous Over-the-Air\\ Federated Edge Learning
}

\author{\IEEEauthorblockN{Saeed Razavikia$^\dagger$, Jaume Anguera Peris$^\dagger$, José Mairton B. da Silva Jr$^{*,\dagger}$, and Carlo Fischione$^\dagger$} 
\IEEEauthorblockA{$^\dagger$School of Electrical Engineering and Computer Science, KTH Royal Institute of Technology, Stockholm, Sweden\\
$^*$Department of Electrical and Computer Engineering, Princeton University, New Jersey, USA\\
Email: $^\dagger$\{sraz, jaumeap, carlofi\}@kth.se, $^*$\{jb8191\}@princeton.edu
}
}

\maketitle

\begin{abstract}
Federated Edge Learning (FEEL) is a distributed machine learning technique where each device contributes to training a global inference model by independently performing local computations with their data. More recently, FEEL has been merged with over-the-air computation (OAC), where the global model is calculated over the air by leveraging the superposition of analog signals. However, when implementing FEEL with OAC, there is the challenge on how to precode the analog signals to overcome any time misalignment at the receiver.

In this work, we propose a novel synchronization-free method to recover the parameters of the global model over the air without requiring any prior information about the time misalignments. For that, we construct a convex optimization based on the norm minimization problem to directly recover the global model by solving a convex semi-definite program.
The performance of the proposed method is evaluated in terms of accuracy and convergence via numerical experiments. We show that our proposed algorithm is close to the ideal synchronized scenario by $10\%$, and performs $4\times$ better than the simple case where no recovering method is used.
\end{abstract}


\section{Introduction}
The popularity of mobile devices and the evolution of sensing and machine learning have led to rapid access to mobile data for training artificial intelligent models. Yet, the variability and the distributed nature of the data do not guarantee that the inference model obtained locally from one device is able to perform well on data from other devices.
To allow for a more collaborative approach, the devices can offload and combine their local data in the cloud, but this solution has been regarded as impractical, mainly due to the communication costs, data privacy concerns, and hardware variability of the devices~\cite{li2020federated}. A more prominent solution is federated learning, in which a centralized node periodically receives and combines the local models from multiple devices to obtain a more accurate global model.

Indeed, the above solutions have gathered interest for federated edge learning (FEEL), which combines federated learning with edge intelligence. Here, the devices periodically offload their models to a server located at the edge of the network, such that these are combined in a federated-learning fashion to obtain a global model that performs well across multiple devices. Specially, FEEL leverages the computational resources at the edge server while minimizing the communication costs at the network~\cite{mcmahan2017communication}.

Yet, the wireless resources located at the edge may become limited as the number of mobile devices participating in FEEL increases or as the dimension of the local gradients increases~\cite{tak2020federated}. To alleviate the communication costs and accelerate the training process, a new paradigm in communication and computation, called over-the-air computation (OAC), concurrently allocates all users over the same frequency resources and leverages the inherent waveform superposition of wireless signals. Specially, by appropriate precoding the analog transmitted signals, it is possible to calculate over the air the class of functions known as nomographic functions~\cite{goldenbaum2014nomographic}, such as the arithmetic mean or the weighted sum. This joint communication-and-computation scheme is particularly useful for FEEL, as the edge server is only interested in constructing a global model from the weighted average of the local models from all devices, so calculating this aggregation over the air using the superposition of analog signals results in much more efficient use of the communication resources and a faster federated-learning system than transmitting the digital signals, decoding them at the receiver, and calculating the aggregation later~\cite{goldenbaum2013harnessing, hellstrom2022wireless}.

However, there are many challenges upon implementing an OAC system. Recent works have made significant progress on this topic regarding the signal processing~\cite{goldenbaum2013harnessing}, the design of the transceivers~\cite{chen2018over}, or the acquisition of channel state information~\cite{ang2019robust}. Nevertheless, there are challenges that are still largely unsolved, such as how to accurately precode the transmitted signals and how to synchronize the clock of all the devices at the level of nanoseconds to maximize the number of participating devices while minimizing the distortions introduced by the channel or the devices themselves. These challenges are particularly important because, in practice, signal precoding is imperfect due to inaccurate channel estimation and non-ideal hardware, and the synchronization across different devices is costly because it requires all devices to time their transmissions accurately based on the propagation time. Therefore, in a realistic scenario, there appears some channel-gain mismatches, time asynchronies, or both~\cite{shao2021bayesian}.

To address these issues, the literature offers different solutions to the misalignment of OAC both for digital and analog systems. Early works~\cite{mahmood2016clock} propose different techniques to achieve clock synchronization using the IEEE 802.11 protocol. These techniques are robust to propagation delays on the channel or within the devices, but they are susceptible to unreliable network transmissions. The work in~\cite{goldenbaum2013robust} implements an analog-modulated system based on coarse block-synchronization to be robust against synchronization errors. Specially, their model is optimal over medium-access strategies that separate the transmissions of nodes in time or code space but not in frequency. The work in~\cite{zhao2021broadband} utilizes digital modulation and channel coding to combat any channel or time misalignments at the receiver, but it still requires decoding the transmitted signals, thus not fully leveraging the communication benefits of analog signals with OAC. The most related work up to date is~\cite{shao2021bayesian}, which tackles the problem of misalignment of analog OAC by constructing matched filters at the receiver, but it assumes that the sampling process knows the exact time misalignments from the different users.

\input{Figs/Fig_FL}

\subsection{Our contribution}
In this work, we focus on the time misalignment of analog OAC and present a novel approach to recovering the arithmetic sum of the transmitted symbols. In particular,

\begin{itemize}
    \item {
        We do not consider any synchronization overhead, such that the receiver has no prior information about the time misalignments of any participating user.
    }
    \item {
        We consider an analog OAC scheme where all devices transmit each parameter of their local model in an uncoded fashion and over a fading, multiple-access wireless channel, such that the receiver sees the superposition of asynchronous signals.
    }
    \item {
        We reformulate the measurement at the receiver in terms of atoms and propose a convex optimization problem to recover the delays and the elements of the gradient, such that we obtain the summation to update the global gradient.
    }
    \item {
        \TODO{We evaluate our proposed algorithm against the MNIST dataset and show how the proposed system achieves competitive results in this task, closing the performance gap down to $10\%$ with respect to the ideal synchronized scenario, and being $4\times$ better than the simple case where no recovery method is implemented.}
    }
\end{itemize}

\subsection{Document organization}
The remainder of this paper is organized as follows. Section \ref{sec:system_model} presents the learning and communication models for the proposed OAC system. In Section \ref{sec:recovery_method}, we present an optimization problem to recover the arithmetic sum of the transmitted symbols. Then, we provide the numerical results in Section \ref{sec:numerical_results}, followed
by the concluding remarks in Section \ref{sec:conclusions}.

\subsection{Notation}
Throughout this paper, scalars are denoted by lower-case letters $x$, and vectors and matrices are denoted by lower-case $\bm{x}$ and upper-case boldface letters $\bm{X}$, respectively. The transpose and Hermitian of a matrix $\bm{X}$ are represented by $\bm{X}^{\mathsf{T}}$ and $\bm{X}^{\mathsf{H}}$, respectively. We further use  $f(t)~\circledast~g(t) $ to show the  convolution between two continuous signals $f(t)$ and $g(t)$. 
For a set $S$, its cardinally is represented  by $|S|$. 
For an integer $N$, $[N]$ stands for $\{1,2,\dots, N\}$. $\bm{X}\succeq \bm{0}$ means that $\bm{X}$ is a positive semidefinite matrix. Finally, the pseudo-inverse of matrix $\bm{X}$ is denoted by $\bm{X}^{\dagger}$, and the element-wise inverse is denoted by $(\cdot)^{\circ -1}$,

\section{System Model}
\label{sec:system_model}
This section is divided into two parts. First, we present the federated learning model, and then we describe the communication model for transmitting the gradients of the inference model in both the uplink and the downlink.

\subsection{Learning model}
Following the same structure as in~\cite{jordan2018communication, mcmahan2017communication, konevcny2016federated, tandon2017gradient, chen2018lag}, consider a federated learning scenario with $K$ devices. For each device $k$, let $\mathcal{D}_k$ denote its local dataset, and let the function $F_k(\bm{w})$ represent the average empirical loss at device $k$ with respect to the model parameters $\bm{w} \in \mathbb{R}^{N}$, with $N$ denoting the number of parameters, such that
\begin{align}
    F_k(\bm{w}) = \frac{1}{|\mathcal{D}_k|} \sum_{j\in \mathcal{D}_k}f_j(\bm{w}),
\end{align}
where $f_j(\bm{w})$ represents the empirical loss function at the data sample $j$ of the $k$-th local dataset $\mathcal{D}_k$. Upon calculating $F_k(\bm{w})$, each device offloads their local model to the edge server for it to construct the global loss function of the model vector $\bm{w}$ from each $\mathcal{D}_k$ as 
\begin{align}
    F(\bm{w}) = \frac{1}{|\mathcal{D}|} \sum_k |\mathcal{D}_k|F_k(\bm{w})
\end{align}
where $\mathcal{D} = \sum_k|\mathcal{D}_k|$ represents the total number of samples from all devices. With that information, the server trains the global model by minimizing the following empirical cost function in a distributed manner 
\begin{align}
   \bm{w}^* =\underset{\bm{w}}{\rm arg min}~F(\bm{w}).
\end{align}

With the idea of preserving privacy, consider further a FEEL framework where each device uses its local dataset $\mathcal{D}_k$ to perform stochastic gradient descent (SGD) to minimize their local loss function $f_j(\bm{w})$. More specifically, let $\bm{w}^{(m)}\in \mathbb{R}^{N}$ and $\bm{\Delta w}_k^{(m)}\in \mathbb{R}^{N}$ represent the parameters of the gradient and the estimate of the gradient of device $k$ at the $m$-th communication round, respectively. Then,
\begin{align}
    \bm{\Delta w}_k^{(m)} = \frac{1}{|\mathcal{D}_k|} \sum_{j\in \mathcal{D}_k}\nabla f_j(\bm{w}^{(m)})
\end{align}
where $\nabla$ denotes the gradient operator. After this computation, the $k$-th device offloads its local model update $\bm{\Delta w}_k^{(m)}$ to the server. Upon receiving the information, the server calculates the global model of the gradient $\bm{\Delta w}^{(m)}$ of the loss function $F(\bm{w})$ from all $K$ users as
\begin{align}
\label{eq:aggreg}
    \bm{\Delta w}^{(m)} = \frac{1}{K}\sum_{k=1}^{K}\bm{\Delta w}_k^{(m)}.
\end{align}
Finally, the server updates the current global model following the gradient descent as 
\begin{align}
    \bm{w}^{(m+1)} = \bm{w}^{(m)} - \eta \bm{\Delta w}^{(m)},
\end{align}
where $\eta$ is the learning rate. The global model is then broadcast back to the devices, and the same procedure is repeated until the model converges to a local minimum. Based on \eqref{eq:aggreg}, we observe that the edge server only needs the aggregation of the local estimations $\bm{\Delta w}_k^{(m)}$, but it does not need the individual gradients from each users. Therefore, this aggregation can be calculated over-the-air following an OAC scheme. Finally, and for simplicity, since the communication procedure is the same for every round, we omit the index $m$ from $\bm{\Delta w}^{(m)}$ and show the gradient at each iteration as $\bm{\Delta w}$.

\subsection{Communication model}
In each communication round, consider all devices to transmit a vector $\bm{x}_{k}\in \mathbb{R}^{N}$ simultaneously to the edge server over a broadband multi-access channel (MAC).
In a typical communication system, since the communications are not ideal because of the characteristics of the receiver and the fact that all devices are not synchronized with each other, let the channel introduce noise, and let a time misalignment appear between the different received signals.
Altogether, we express the received signal at the edge server at time $t$ as
\begin{align}
    \bm{y}(t) = \sum_{k=1}^{K}h_k\bm{x}_k(t-\tau_k) + \bm{z}(t).
    \label{eq:receivedSignal}
\end{align}
The coefficients $h_k$ originate from transmitting the symbols from the $k$-th device to the edge server over a fading channel, which are assumed to be Rayleigh distributed and known at the transmitter. The parameters $\tau_k$ represent the unknown delays of $k$-th device, and are considered to take arbitrary continuous values in $[0,T]$. Finally, the additive term $\bm{z}(t)$ represents the white Gaussian noise (AWGN), which is considered to be distributed according to $\mathcal{N}(\bm{0},\sigma_z^2\bm{I})$.

Considering this uplink transmission model, we are interested in designing a transmitter and a receiver that reverts the effect of the channel and allows all $K$ devices to transmit all $N$ parameters from $\bm{\Delta w}_k$ at once, such that the receiver recovers \eqref{eq:aggreg} from the received signals. For the resource allocation, consider $\Bar{B}$ to be the total available bandwidth in our system, and let $B = \Bar{B}/N$ be the bandwidth allocated to simultaneously transmit one of the elements of the gradient $\bm{\Delta w}_k$ for all devices $k\in[K]$\footnote{For large $N$, the whole uplink transmission can be separated in different batches, such that all $N$ parameters are transmitted to the server over different time periods.}. Then, consider that we construct $N$ orthogonal band-limited signals $r_i(t)$\footnote{The waveform $r(t)$ can be constructed over the band $[-B/2,B/2]$ using the sinc kernel ${\rm sinc} (Bt)$. Specially, since the sinc kernel decays relatively fast, $r(t)$ would be approximately supported on the interval $[-3T/2,3T/2]$.} for $i\in[N]$ to simultaneously transmit the $i$-th element of the gradient ${\Delta w}_{i,k} := [\bm{\Delta w}]_{i,k}$ of all devices $k \in [K]$ over the $i$-th frequency band, with
$$R_i(f) = \mathcal{F}\{r_i(t)\} = 0, \quad \forall |f|>B/2,$$
where $\mathcal{F}\{\cdot\}$ denotes the Fourier transform.

At the transmitter side, let each device $k$ construct a positive representation of the gradient ${\Delta w}_{i,k}$ as
\begin{equation*}
    {\Delta w}_{i,k}^{+} := {\Delta w}_{i,k}  + \gamma \geq 0.
\end{equation*}
where $\gamma$ is selected sufficiently large to satisfy the inequality above\footnote{The parameter $\gamma$ can be selected before initiating the FEEL process and be known at the receiver, so it suffices to make $\gamma$ large enough to satisfy the inequality, with the only condition that it needs to be larger than the possible minimum of all gradients, i.e., $\gamma \geq \min_{i,k}{\Delta w}_{i,k}$. This minimum can also be shared by the edge server, in which case the overhead of transmitting one scalar per device is negligible}. With that, each device encodes the information of the gradient\footnote{This encoding step also includes quantization, so the encoded information in the transmitted signal loses accuracy. Here we do not consider this loss in accuracy, but one could refer to \cite{zhu2020one} for more information about the effect of the quantization error on the inference problem.} into the signal $\bm{x}_k(t) = \bm{\Delta w}_{k}^{+} \frac{h^*_k}{|h_k|}\sqrt{p_k} \bm{r}(t)$ from the known Rayleigh fading coefficients $h_k$ and the $N\times1$ vector signal $\bm{r}(t)$ whose elements $r_i(t)$ satisfy $r_i(t) \circledast r_j(t) = 0$ for any $i\neq j$. Here we assume that all devices compensate for the effect of the fading $h_k$ by setting $p_k = |h_k|^{-2}$; otherwise it would result in weaker signal strength.

\input{Figs/SamplingModel}

The task of the receiver is then to estimate the global gradient $\bm{\Delta w}^{+}$ from the received signal \eqref{eq:receivedSignal}. From the description above, the $i$-th element of the received signal ${y}_{i}(t) := [\bm{y}]_i(t)$ can be expressed as
\begin{align}
    \label{eq:noiseless}
      {y}_{i}(t) =\sum_{k=1}^{K}{\Delta w}_{i,k}^{+} \, {r}_i(t) + z_{i}(t),\quad \forall i \in [N].
\end{align}
After the signal goes through a matched filter, we obtain
\begin{align}
    \tilde{y}_{i}(t) & = y_{i}(t) \circledast r_i^*(-t) \nonumber \\
    & = \Big(\sum_{k=1}^{K}{\Delta w}_{i,k}^{+} \, {r}_i(t-\tau_k) + z_{i}(t)\Big) \circledast r_i^*(-t) \nonumber \\
    & = \sum_{k=1}^{K}{\Delta w}_{i,k}^{+} \, g_i(t-\tau_k) +  \tilde{z}_{i}(t), \label{eq:CorrSamples}
\end{align}
where $g(t)$ represents the convoluted waveform at time $t$.
Since $\tilde{y}_{i}(t)$ is band-limited and approximately time-limited, the receiver samples $\tilde{y}_{i}(t)$ at rate $1/B$ in the interval $[-T/2,T/2]$ to collect all its degree of freedom  $L=BT=:2M+1$ samples.  
If we mathematically manipulate \eqref{eq:CorrSamples} by applying the discrete Fourier transform (DFT) and the inverse DFT, the $\ell$-th sample of the signal $\tilde{y}_{i}(t)$, denoted by $Y_{i}(\ell)$, in the interval $\frac{\ell}{B} \in [-T/2,T/2]$ can be expressed by~\cite{bayat2020separating}
\begin{align} \label{eq:FreqMeas}
 Y_{i}(\ell) = \sum_{k=1}^K{\Delta w}_{i,k}^{+}\; \theta_{i,k}(\ell) + Z_{i}(\ell),
 \end{align}
where $Z_{i}(\ell) = \tilde{z}_i(\tfrac{\ell}{B})$,
\begin{equation}
    \theta_{i,k}(\ell) = \sum_{q=-M}^{M} [\bm{a}(\tau_k)]_q \; g_i{(\tfrac{\ell}{B})}\,e^{\frac{j2\pi\ell q}{L}},
\end{equation}
and 
\begin{equation}
    [\bm{a}(\tau_k)]_q = \frac{1}{L}\sum_{r=-M}^{M}e^{\tfrac{j2\pi(q-L\tau_{k})r}{L}},
\end{equation}
with $\bm{a}(\tau)\in \mathbb{C}^{L}$ being the atom or building block~\cite{chandrasekaran2012convex}. Without loss of generality, we assume all $g_i(t)$ to be periodic with duration $T$\footnote{
We assume periodicity to obtain \eqref{eq:FreqMeas}. Clearly, this assumption is not practical because the waveform signal is not time-limited. However, we can consider the quasi-periodic waveform satisfying $g(\frac{\ell}{B}) = 0$ for $\ell \notin [-\lfloor L/2 \rfloor,\lfloor L/2 \rfloor]$.}, $T_{\rm max} = 1$, and $\tau_k \in [0,1)$ for $k\in [K]$. We further map the frequency indices $\ell$ from $-\lfloor B/2 \rfloor,\ldots,0,\ldots, \lfloor B/2 \rfloor$ to $0,\dots,L-1$, where $L= 2\lfloor B/2 \rfloor+1$ is the total number of frequency samples. Furthermore, to alleviate the notation, we rewrite \eqref{eq:FreqMeas} as 
\begin{align}\label{eq:Obervation}
    \bm{Y}_i = \bm{G}_i\bm{X}_i + \bm{Z}_i,
\end{align}
where $\bm{X}_i = \sum_{k=1}^K{\Delta w}_{i,k}^{+} \, \bm{a}(\tau_k)\in \mathbb{C}^{L}$, $\bm{G}_{i} \in \mathbb{C}^{L\times L}$ is the waveform matrix whose $(\ell,q)$-th element is defined as $[\bm{G}_i]_{(\ell,q)} = g_i(\tfrac{\ell}{B})e^{\frac{j2\pi\ell q}{L}}$, $\bm{Z}_i = [Z_i(1),\ldots,Z_i(L)]^{\mathsf{T}}$, and $\bm{Y}_i = [Y_i(1),\ldots,Y_i(L)]^{\mathsf{T}}$.  Altogether, the sampling scheme is depicted in Fig.~\ref{fig:SamplingScheme}. Since the waveform $r_i(t)$ is known, we can remove the effect of $g_{i}(t)$ in \eqref{eq:Obervation} by multiplying the measurements to the Fourier transform inverse $(\bm{G}_{i})^{-1}$ (assuming $\bm{G}_{i}$ is invertible for any $i \in [N]$). Then, we observe 
\begin{align}
    \label{eq:Measurment}
    \bm{V}_{i} = \bm{X}_{i} +  \hat{\bm{Z}}_{i}, \quad \forall i~\in[N],
\end{align}
where $\bm{V}_{i} = (\bm{G}_{i})^{-1} \bm{Y}_{i}$, and $\hat{\bm{Z}}_{i}$ is the AWGN additive noise with bounded variance $\sigma_z^2$. Considering this, the next section presents a method to perfectly recover $\sum_{k=1}^{K}{\Delta w}_{i,k}$ from $\bm{V}_i$.

\section{Recovery Method}
\label{sec:recovery_method}
The problem presented in Section \ref{sec:system_model} and summarized in \eqref{eq:Measurment} is closely related to the classical problem of \textit{line spectrum estimation} or \textit{super resolution} problem~\cite{candes2014towards}, in which we seek to estimate all the delays and the amplitudes $\{\tau_k,{\Delta w}_{i,k}^{+} \}_{k=1}^K$ from a mixture of sinusoids. The main difference is that the edge server is interested in the mean $\tfrac{1}{K}\sum_{k=1}^{K}{\Delta w}_{i,k}^{+}$ instead of each individual component ${\Delta w}_{i,k}^{+}$. Our proposed solution is to estimate directly the mean of gradients $s_{i}:=\tfrac{1}{K}\sum_{k=1}^{K}{\Delta w}_{i,k}^{+}$ from $\eqref{eq:Measurment}$ using the notion of atomic norm of the signal $\bm{X}_{i}$.

To begin, define the atomic set as
\begin{equation}
    \mathcal{A} := \{\bm{a}(\tau): \tau \in [0,1)\},
\end{equation}
and the associated Minkowski functional over the set $\mathcal{A}$ as~\cite{chandrasekaran2012convex}
\begin{align}
\|\bm{X}\|_{\mathcal{A}}:=\inf_{\substack{\beta_j\in\mathbb{R}^{+}\\\tau_j\in[0,1)}}\Big\{\sum_{j}\beta_j~:~\bm{X}=\sum_{j}\beta_{j} \bm{a}(\tau_j)\Big\}.
    \label{eq:atomicNorm}
\end{align}
Finding the optimal parameters in \eqref{eq:atomicNorm} is not an easy task because it involves an infinite-dimensional variable optimization due to the continuity of the set $\mathcal{A}$. Alternatively, we can  rewrite \eqref{eq:atomicNorm} using the Carathéodory-Fejér-Pisarenko decomposition~\cite{georgiou2007caratheodory} as the following semi-definite program
 \begin{align}
 &\|\bm{X}\|_{\mathcal{A}}=\min_{\bm{u}\in\mathbb{C}^{L},t>0}{\;\frac{1}{2}}(\mathsf{Re}(u_1) +t)\nonumber\\[.2cm]
 &{\rm s.t.}~~\begin{bmatrix}
 {\rm Toep}(\bm{u})&\bm{F}^{-\mathsf{H}}\bm{X}\\
 \bm{X}^{\mathsf{H}}\bm{F}^{-1}& t
 \end{bmatrix}\succeq \bm{0},
 \end{align}
where $\bm{F}$ denotes the DFT matrix whose elements are given by $[\bm{F}]_{(k,l)}:=\frac{1}{L} {\rm e}^{-{\rm j} 2\pi lk/L}$, and the Hermitian Toeplitz matrix ${\rm Toep}(\bm{u})$ for a vector $\bm{u}\in \mathbb{C}^{L}$, is defined as
\begin{align}
    {\rm Toep}(\bm{u}) := \left[ \begin{matrix}
u_{1} & u_{2}  & \dots& u_{L} \\
u_{2}^* & u_{1}  & \dots& u_{L-1} \\
\vdots & \vdots & \ddots&  \vdots\\
u_{L}^* & u_{L-1}^*  & \dots & u_{1} 
\end{matrix}  \right]. 
\end{align}
Hence, the optimization problem to jointly recover the delays $\tau_k$ and the gradients ${\Delta w}_{i,k}^{+}$ can be obtained by searching for signals $\bm{X}_i$ that are both sparse in the continuous atom set $\mathcal{A}$ (small atom norms) and close to the observation $\bm{V}_{i}$, i.e.,
\begin{align}
\label{eq:AtomicLasso}
    \min_{\bm{X}_i} \|\bm{X}_i\|_{\mathcal{A}} + \lambda \|\bm{X}_i-\bm{V}_{i} \|_2^2.
\end{align}
Here, the regularization parameter $\lambda > 0$ needs to be chosen appropriately. By employing the atomic decomposition, \eqref{eq:AtomicLasso} can be reformulated as follows
 \begin{align}\label{eq:atomicnormminize}
 &\hat{\bm{X}}_{i}= \min_{\substack{\bm{x},t>0\\\bm{u}\in\mathbb{C}^{L}}} \; \frac{1}{2}(\mathsf{Re}(u_1) +t) +  \lambda \|\bm{X}_i-\bm{V}_{i} \|_2^2\nonumber\\[.2cm]
 &{\rm s.t.}~~\begin{bmatrix}
 {\rm Toep}(\bm{u})&\bm{F}^{-\mathsf{H}}\bm{X}_i\\
 \bm{X}_i^H\bm{F}^{-1}& t
 \end{bmatrix}\succeq \bm{0}.
 \end{align}
This convex optimization problem can be solved efficiently using standard tools from convex optimization~\cite{boyd2004convex}.

After obtaining the Toeplitz matrix ${\rm Teop}(\bm{u})$, we employ the Vandermonde decomposition via solving a generalized eigenvalue problem~\cite{hua1990matrix} as
\begin{equation*}
    {\rm Toep}(\bm{u}) = \sum_{k=1}^{K'} {\Delta w}_{i,k}^{+'}\bm{a}(\tau_k^{'})\bm{a}(\tau_k^{'})^{\mathsf{H}}
\end{equation*}
to identify the support $\{\tau_k^{'}\}_{k=1}^{K'}$ as well as the atomic norm $\|\hat{\bm{X}}_i\|_{\mathcal{A}} = \sum_{k=1}^{K} {\Delta w'}_{i,k}^{+}$. With that, the amplitudes, and subsequently the mean value $\hat{s}_{i} = \tfrac{1}{K}\|\hat{\bm{X}}_i\|_{\mathcal{A}} = \tfrac{1}{K}\sum_{k=1}^{K} {\Delta w}_{i,k}^{+} - \gamma$, are obtained. Finally, the average gradient is estimated from $ \widehat{\bm{\Delta w}}  = [ \hat{s}_{1},\ldots,\hat{s}_{N}  ]^{\mathsf{T}} $.
\begin{rem}
    It is established that problem \eqref{eq:atomicnormminize} can recover the positive  mean value  $s_i$ without requiring any separation between the delays \cite{morgenshtern2016super}, as long as $K\leq \lfloor \frac{L-1}{2}\rfloor $, and
    the estimated parameters $\hat{s}_i$ satisfy 
    \begin{align}
        |s_i - \hat{s}_i| = 
        \mathcal{O}\bigg(\sigma_z \sqrt{\frac{\log{L}}{L} }\bigg), \quad \forall i \in [N]
    \end{align}
    where $\sigma_z$ denotes the variance of noise and $L$ is the number of samples.
    \label{rem:separationSamples}
\end{rem}

\section{Numerical Results}
\label{sec:numerical_results}
This section examines the performance of our proposed algorithm in terms of the normalized mean squared error (NMSE) between the true sum of models and the estimated one for different numbers of users $K$ and Fourier samples $L$. We also evaluate our method against the MNIST dataset and show how the proposed system can achieve competitive results in this task while using a simple recovery method. In all the experiments, the channel coefficients $\{h_k\}_{k=1}^K$ are \textit{i.i.d} generated uniformly on the unit sphere, and the delays $\{\tau_k\}_{k=1}^K$ are uniformly distributed at random between $0$ and $1$, i.e., $\tau_k\sim \mathcal{U}[0,1)$. Moreover, all waveforms $r_i(t)$ are randomly generated using $\sum_{\ell = -L -M}^{L+M} a_{i} {\rm sinc(tB - \ell)}$ with  $a_{i} \sim \mathcal{N}(0,1/L)$ for all $i\in [N]$. The optimization problem in \eqref{eq:atomicnormminize} is implemented using the SDPT3 package of CVX in MATLAB.

\input{Figs/Fig_KvsSNR}

For the first experiment, we check the performance of the proposed algorithm for estimating the summation of each individual element of the gradient vector ${\Delta w}_{i}$ for different signal to noise ratio (SNR), defined as ${\rm SNR} = 20\log{(\tfrac{\|\bm{G} \bm{X}\|_2}{\|\bm{Z}\|_2})}$. Figure \ref{fig:samples-Users_NMSE} (a) shows the NMSE for $K=5$ users over different SNRs within $[4,20]$ dB and over different number of Fourier samples $L=\{8,16,32,64,128\}$. As we can observe, the error decreases for increasing SNR and for increasing number of samples. We further repeat the simulation to check the effect of the number of users on the recovery problem. The results are shown in Figure \ref{fig:samples-Users_NMSE} (b) for $L=128$ samples. As the number of users increases, the NMSE error also increases, which shows the need for more  samples. Specially, the error aggravates for a large number of users, e.g., $K\geq 40$. This comes from the fact that having more users results in two or more parameters being close to each other, thus making it more difficult to satisfy the conditions from Remark \ref{rem:separationSamples}. Consequently, the optimization in \eqref{eq:atomicnormminize} cannot perfectly separate the amplitude and delays of each individual device $k$, resulting in a worse NMSE.

For the last experiment, we evaluate the performance of the estimation for the FEEL problem for $K=10$ devices when the task is a multi-label classification on the MNIST dataset~\cite{lecun1998gradient}. We used a simple neural network that consists of an input layer of $784$ nodes, a hidden layer of $100$ nodes, and an output layer of $10$ nodes. We compare the performance of our proposed method with the standard method when the communication system uses ideal versus imperfect synchronization with $\text{SNR} = 5$ dB. Figure \ref{fig:MNIST} depicts the accuracy for these cases with different number of  samples $L=128$ and $L=256$. \TODO{Note that the accuracy of our proposed algorithm increases for increasing number of samples, closing the gap with respect to the ideal synchronized scenario down to $10\%$, and being up to $4$ times better than just recovering the global model directly from the received data without doing any processing. Overall, the results suggest that our proposed method can successfully learn the global model even in a blind imperfect synchronization scenario.}

\input{Figs/Fig_MNIST}

\section{Conclusions}
\label{sec:conclusions}
In this paper, we consider the FEEL problem where each device contributes to training a global inference model by independently performing local computations with their data. We explained that typical over-the-air computation methods usually make the idealistic assumption that there is perfect synchronization between the devices and the receiver. Other existing methods solve this misalignment problem, but they assume the delays to be known. Here, we instead focus on the OAC problem considering that there is no prior information on channel delays over the AWGN channel. We proposed a novel synchronization-free method to recover the global inference model without requiring any prior information about the devices' delays. For that, we developed an atomic norm minimization problem in order to recover the summation of the gradient by solving a convex semi-definite program. \TODO{Finally, we evaluated the performance of the recovery method in terms of accuracy and convergence via numerical experiments, and showed that our proposed system is close to the ideal synchronized scenario by $10\%$, and performs $4\times$ better than using no recovery methods.}

\vspace{-2pt}
\bibliographystyle{ieeetr}
\bibliography{IEEEabrv,Ref2}


\end{document}

%% file: Figs/Fig_FL.tex
\begin{figure*}[!t]
    \centering
    
\scalebox{0.9}{
\tikzset{every picture/.style={line width=0.75pt}} 

\begin{tikzpicture}[x=0.75pt,y=0.75pt,yscale=-1,xscale=1]

\draw  [fill={rgb, 255:red, 40; green, 33; blue, 33 }  ,fill opacity=1 ] (553.21,114.23) .. controls (553.21,109.93) and (562.06,106.45) .. (572.99,106.45) .. controls (583.91,106.45) and (592.77,109.93) .. (592.77,114.23) .. controls (592.77,118.52) and (583.91,122) .. (572.99,122) .. controls (562.06,122) and (553.21,118.52) .. (553.21,114.23) -- cycle ;
\draw  [color={rgb, 255:red, 255; green, 252; blue, 252 }  ,draw opacity=1 ][fill={rgb, 255:red, 247; green, 246; blue, 246 }  ,fill opacity=1 ] (553.21,110.34) .. controls (553.21,106.05) and (561.92,102.57) .. (572.66,102.57) .. controls (583.4,102.57) and (592.11,106.05) .. (592.11,110.34) .. controls (592.11,114.63) and (583.4,118.11) .. (572.66,118.11) .. controls (561.92,118.11) and (553.21,114.63) .. (553.21,110.34) -- cycle ;
\draw  [fill={rgb, 255:red, 40; green, 33; blue, 33 }  ,fill opacity=1 ] (553.21,106.45) .. controls (553.21,102.16) and (562.06,98.68) .. (572.99,98.68) .. controls (583.91,98.68) and (592.77,102.16) .. (592.77,106.45) .. controls (592.77,110.75) and (583.91,114.23) .. (572.99,114.23) .. controls (562.06,114.23) and (553.21,110.75) .. (553.21,106.45) -- cycle ;
\draw  [color={rgb, 255:red, 255; green, 252; blue, 252 }  ,draw opacity=1 ][fill={rgb, 255:red, 247; green, 246; blue, 246 }  ,fill opacity=1 ] (553.21,102.57) .. controls (553.21,98.28) and (562.06,94.8) .. (572.99,94.8) .. controls (583.91,94.8) and (592.77,98.28) .. (592.77,102.57) .. controls (592.77,106.86) and (583.91,110.34) .. (572.99,110.34) .. controls (562.06,110.34) and (553.21,106.86) .. (553.21,102.57) -- cycle ;
\draw  [fill={rgb, 255:red, 40; green, 33; blue, 33 }  ,fill opacity=1 ] (553.21,98.68) .. controls (553.21,94.39) and (562.06,90.91) .. (572.99,90.91) .. controls (583.91,90.91) and (592.77,94.39) .. (592.77,98.68) .. controls (592.77,102.97) and (583.91,106.45) .. (572.99,106.45) .. controls (562.06,106.45) and (553.21,102.97) .. (553.21,98.68) -- cycle ;
\draw  [color={rgb, 255:red, 255; green, 252; blue, 252 }  ,draw opacity=1 ][fill={rgb, 255:red, 247; green, 246; blue, 246 }  ,fill opacity=1 ] (553.21,94.8) .. controls (553.21,90.5) and (562.21,87.02) .. (573.32,87.02) .. controls (584.43,87.02) and (593.43,90.5) .. (593.43,94.8) .. controls (593.43,99.09) and (584.43,102.57) .. (573.32,102.57) .. controls (562.21,102.57) and (553.21,99.09) .. (553.21,94.8) -- cycle ;
\draw  [fill={rgb, 255:red, 40; green, 33; blue, 33 }  ,fill opacity=1 ] (553.21,90.91) .. controls (553.21,86.62) and (562.06,83.14) .. (572.99,83.14) .. controls (583.91,83.14) and (592.77,86.62) .. (592.77,90.91) .. controls (592.77,95.2) and (583.91,98.68) .. (572.99,98.68) .. controls (562.06,98.68) and (553.21,95.2) .. (553.21,90.91) -- cycle ;

\draw    (552,103) -- (520,103) ;
\draw [shift={(520,103)}, rotate = 360] [fill={rgb, 255:red, 0; green, 0; blue, 0 }  ][line width=0.08]  [draw opacity=0] (3.57,-1.72) -- (0,0) -- (3.57,1.72) -- cycle    ;
\draw    (405,103) -- (387,103) ;
\draw [shift={(387,103)}, rotate = 360] [fill={rgb, 255:red, 0; green, 0; blue, 0 }  ][line width=0.08]  [draw opacity=0] (3.57,-1.72) -- (0,0) -- (3.57,1.72) -- cycle    ;
\draw    (298,103) -- (271,103) ;
\draw [shift={(271,103)}, rotate = 360] [fill={rgb, 255:red, 0; green, 0; blue, 0 }  ][line width=0.08]  [draw opacity=0] (3.57,-1.72) -- (0,0) -- (3.57,1.72) -- cycle    ;
\draw    (534,57) -- (534,99) ;
\draw [shift={(534,102)}, rotate = 270] [fill={rgb, 255:red, 0; green, 0; blue, 0 }  ][line width=0.08]  [draw opacity=0] (3.57,-1.72) -- (0,0) -- (3.57,1.72) -- cycle    ;
\draw    (480,57) -- (534,57) ;
\draw   (262.96,84.97) -- (256.11,76) -- (270,76.12) -- cycle ;
\draw    (262.96,84.97) -- (263,105) ;

\draw  [fill={rgb, 255:red, 40; green, 33; blue, 33 }  ,fill opacity=1 ] (554.21,251.23) .. controls (554.21,246.93) and (563.06,243.45) .. (573.99,243.45) .. controls (584.91,243.45) and (593.77,246.93) .. (593.77,251.23) .. controls (593.77,255.52) and (584.91,259) .. (573.99,259) .. controls (563.06,259) and (554.21,255.52) .. (554.21,251.23) -- cycle ;
\draw  [color={rgb, 255:red, 255; green, 252; blue, 252 }  ,draw opacity=1 ][fill={rgb, 255:red, 247; green, 246; blue, 246 }  ,fill opacity=1 ] (554.21,247.34) .. controls (554.21,243.05) and (562.92,239.57) .. (573.66,239.57) .. controls (584.4,239.57) and (593.11,243.05) .. (593.11,247.34) .. controls (593.11,251.63) and (584.4,255.11) .. (573.66,255.11) .. controls (562.92,255.11) and (554.21,251.63) .. (554.21,247.34) -- cycle ;
\draw  [fill={rgb, 255:red, 40; green, 33; blue, 33 }  ,fill opacity=1 ] (554.21,243.45) .. controls (554.21,239.16) and (563.06,235.68) .. (573.99,235.68) .. controls (584.91,235.68) and (593.77,239.16) .. (593.77,243.45) .. controls (593.77,247.75) and (584.91,251.23) .. (573.99,251.23) .. controls (563.06,251.23) and (554.21,247.75) .. (554.21,243.45) -- cycle ;
\draw  [color={rgb, 255:red, 255; green, 252; blue, 252 }  ,draw opacity=1 ][fill={rgb, 255:red, 247; green, 246; blue, 246 }  ,fill opacity=1 ] (554.21,239.57) .. controls (554.21,235.28) and (563.06,231.8) .. (573.99,231.8) .. controls (584.91,231.8) and (593.77,235.28) .. (593.77,239.57) .. controls (593.77,243.86) and (584.91,247.34) .. (573.99,247.34) .. controls (563.06,247.34) and (554.21,243.86) .. (554.21,239.57) -- cycle ;
\draw  [fill={rgb, 255:red, 40; green, 33; blue, 33 }  ,fill opacity=1 ] (554.21,235.68) .. controls (554.21,231.39) and (563.06,227.91) .. (573.99,227.91) .. controls (584.91,227.91) and (593.77,231.39) .. (593.77,235.68) .. controls (593.77,239.97) and (584.91,243.45) .. (573.99,243.45) .. controls (563.06,243.45) and (554.21,239.97) .. (554.21,235.68) -- cycle ;
\draw  [color={rgb, 255:red, 255; green, 252; blue, 252 }  ,draw opacity=1 ][fill={rgb, 255:red, 247; green, 246; blue, 246 }  ,fill opacity=1 ] (554.21,231.8) .. controls (554.21,227.5) and (563.21,224.02) .. (574.32,224.02) .. controls (585.43,224.02) and (594.43,227.5) .. (594.43,231.8) .. controls (594.43,236.09) and (585.43,239.57) .. (574.32,239.57) .. controls (563.21,239.57) and (554.21,236.09) .. (554.21,231.8) -- cycle ;
\draw  [fill={rgb, 255:red, 40; green, 33; blue, 33 }  ,fill opacity=1 ] (554.21,227.91) .. controls (554.21,223.62) and (563.06,220.14) .. (573.99,220.14) .. controls (584.91,220.14) and (593.77,223.62) .. (593.77,227.91) .. controls (593.77,232.2) and (584.91,235.68) .. (573.99,235.68) .. controls (563.06,235.68) and (554.21,232.2) .. (554.21,227.91) -- cycle ;

\draw    (553,239) -- (520,239) ;
\draw [shift={(520,239)}, rotate = 360] [fill={rgb, 255:red, 0; green, 0; blue, 0 }  ][line width=0.08]  [draw opacity=0] (3.57,-1.72) -- (0,0) -- (3.57,1.72) -- cycle    ;
\draw    (405,240) -- (387,240) ;
\draw [shift={(387,240)}, rotate = 0.5] [fill={rgb, 255:red, 0; green, 0; blue, 0 }  ][line width=0.08]  [draw opacity=0] (3.57,-1.72) -- (0,0) -- (3.57,1.72) -- cycle    ;
\draw    (298,240) -- (275,240) ;
\draw [shift={(272,240)}, rotate = 0.38] [fill={rgb, 255:red, 0; green, 0; blue, 0 }  ][line width=0.08]  [draw opacity=0] (3.57,-1.72) -- (0,0) -- (3.57,1.72) -- cycle    ;
\draw    (535,195) -- (535,236) ;
\draw [shift={(535,239)}, rotate = 270] [fill={rgb, 255:red, 0; green, 0; blue, 0 }  ][line width=0.08]  [draw opacity=0] (3.57,-1.72) -- (0,0) -- (3.57,1.72) -- cycle    ;
\draw    (480,195) -- (535,195) ;
\draw   (263.96,221.97) -- (257.11,213) -- (271,213.12) -- cycle ;
\draw    (263.96,221.97) -- (264,242) ;

\draw [line width=1.5]    (158.72,122.99) -- (140,182.45) ;
\draw [line width=1.5]    (158.72,122.99) -- (161.39,185) ;
\draw [line width=1.5]    (158.72,122.99) -- (181,178.2) ;
\draw    (160.05,153.99) -- (169.86,150.6) ;
\draw    (173.87,159.8) -- (160.5,165.46) ;
\draw    (165.85,140.26) -- (159.61,142.53) ;
\draw    (177.43,169.14) -- (161.39,175.66) ;
\draw    (161.39,175.66) -- (143.57,169.99) ;
\draw    (160.5,165.46) -- (146.24,161.5) ;
\draw    (149.36,152.72) -- (160.05,153.99) ;
\draw    (152.48,141.11) -- (159.61,142.53) ;
\draw  [draw opacity=0][fill={rgb, 255:red, 156; green, 133; blue, 133 }  ,fill opacity=1 ] (145.35,106) -- (172.09,106) -- (172.09,123.83) -- (145.35,123.83) -- cycle ; \draw   (149.8,106) -- (149.8,123.83)(154.26,106) -- (154.26,123.83)(158.72,106) -- (158.72,123.83)(163.17,106) -- (163.17,123.83)(167.63,106) -- (167.63,123.83) ; \draw   (145.35,110.46) -- (172.09,110.46)(145.35,114.91) -- (172.09,114.91)(145.35,119.37) -- (172.09,119.37) ; \draw   (145.35,106) -- (172.09,106) -- (172.09,123.83) -- (145.35,123.83) -- cycle ;

\draw    (251,98) -- (191.45,140.26) ;
\draw [shift={(189,142)}, rotate = 324.64] [fill={rgb, 255:red, 0; green, 0; blue, 0 }  ][line width=0.08]  [draw opacity=0] (3.57,-1.72) -- (0,0) -- (3.57,1.72) -- cycle    ;
\draw [line width=0.75]  [dash pattern={on 0.84pt off 2.51pt}]  (252.55,105.74) -- (226,124.58) -- (193,148) ;
\draw [shift={(255,104)}, rotate = 144.64] [fill={rgb, 255:red, 0; green, 0; blue, 0 }  ][line width=0.08]  [draw opacity=0] (3.57,-1.72) -- (0,0) -- (3.57,1.72) -- cycle    ;
\draw    (254.91,223.56) -- (195.93,180.51) ;
\draw [shift={(193.51,178.74)}, rotate = 36.13] [fill={rgb, 255:red, 0; green, 0; blue, 0 }  ][line width=0.08]  [draw opacity=0] (3.57,-1.72) -- (0,0) -- (3.57,1.72) -- cycle    ;
\draw [line width=0.75]  [dash pattern={on 0.84pt off 2.51pt}]  (248.07,227.49) -- (221.77,208.29) -- (189.09,184.44) ;
\draw [shift={(250.49,229.26)}, rotate = 216.13] [fill={rgb, 255:red, 0; green, 0; blue, 0 }  ][line width=0.08]  [draw opacity=0] (3.57,-1.72) -- (0,0) -- (3.57,1.72) -- cycle    ;

\draw    (286,57) -- (328,57) ;
\draw [shift={(328,57)}, rotate = 180] [fill={rgb, 255:red, 0; green, 0; blue, 0 }  ][line width=0.08]  [draw opacity=0] (3.57,-1.72) -- (0,0) -- (3.57,1.72) -- cycle    ;
\draw    (286,57) -- (270,71) ;
\draw    (80,164) -- (80,207) ;
\draw    (80,207) -- (154,207) ;
\draw    (154,190) -- (154,207) ;
\draw [shift={(154,187)}, rotate = 90] [fill={rgb, 255:red, 0; green, 0; blue, 0 }  ][line width=0.08]  [draw opacity=0] (3.57,-1.72) -- (0,0) -- (3.57,1.72) -- cycle    ;
\draw    (287,195) -- (328,195) ;
\draw [shift={(328,195)}, rotate = 180] [fill={rgb, 255:red, 0; green, 0; blue, 0 }  ][line width=0.08]  [draw opacity=0] (3.57,-1.72) -- (0,0) -- (3.57,1.72) -- cycle    ;
\draw    (287,195) -- (271,208) ;

\draw    (405,89) -- (520,89) -- (520,115) -- (405,115) -- cycle  ;
\draw (407,93.4) node [anchor=north west][inner sep=0.75pt]  [font=\footnotesize]  {$\bm{\Delta w}_{1}^{(m)} = \nabla F_{1}(\bm{w}^{(m)})$};
\draw    (298,88) -- (387,88) -- (387,117) -- (298,117) -- cycle  ;
\draw (300,92) node [anchor=north west][inner sep=0.75pt]  [font=\footnotesize]  {$\bm{x}_1=\bm{r}(t) \bm{\Delta w}_{1}^{(m)}$};
\draw    (328,47) -- (480,47) -- (480,69) -- (328,69) -- cycle  ;
\draw (331,51.4) node [anchor=north west][inner sep=0.75pt]  [font=\footnotesize]  {$\bm{w}^{(m)}\hspace{-1pt} = \hspace{-1pt}\bm{w}^{(m-1)}\hspace{-1pt} -\eta \bm{\Delta w}^{(m-1)}$};
\draw (545,60) node [anchor=north west][inner sep=0.75pt]   [align=left] {{\footnotesize Local Dataset}};
\draw    (405,225) -- (520,225) -- (520,251) -- (405,251) -- cycle  ;
\draw (407,229) node [anchor=north west][inner sep=0.75pt]  [font=\footnotesize]  {$\bm{\Delta w}_{K}^{(m)}  = \nabla F_{K}\hspace{-1.5pt}(w^{(m)})$};
\draw    (298,224) -- (387,224) -- (387,252) -- (298,252) -- cycle  ;
\draw (302,228.4) node [anchor=north west][inner sep=0.75pt]  [font=\footnotesize]  {$\hspace{-2pt}\bm{x}_K\hspace{-2pt}= \bm{r}(t) \bm{\Delta w}_{K}^{(m)}$};
\draw    (328,184) -- (480,184) -- (480,206) -- (328,206) -- cycle  ;
\draw (332,188.4) node [anchor=north west][inner sep=0.75pt]  [font=\footnotesize]  {$\bm{w}^{(m)}\hspace{-1pt} =\hspace{-1pt}\bm{w}^{(m-1)} \hspace{-2pt}-\eta {\bm{\Delta w}}^{(m-1)}$};
\draw (545,196) node [anchor=north west][inner sep=0.75pt]   [align=left] {{\footnotesize Local Dataset}};
\draw (390,132) node [anchor=north west][inner sep=0.75pt]  [font=\LARGE]  {$\vdots $};
\draw    (10,136) -- (138,136) -- (138,158) -- (10,158) -- cycle  ;
\draw (77,147) node  [font=\footnotesize]  {$\hspace{-3pt}\bm{\Delta w}^{(m)}= \sum_{k}\bm{\Delta w}_k^{(m)}/K$};
\draw (115,220) node  [font=\footnotesize]  {$ \bm{w}^{(m+1)} = \bm{w}^{(m)} - \eta \bm{\Delta w}^{(m)}$};

\end{tikzpicture}
}
    \caption{Diagram of federated edge learning with over-the-air computation }
    \label{fig:federated}
\end{figure*}
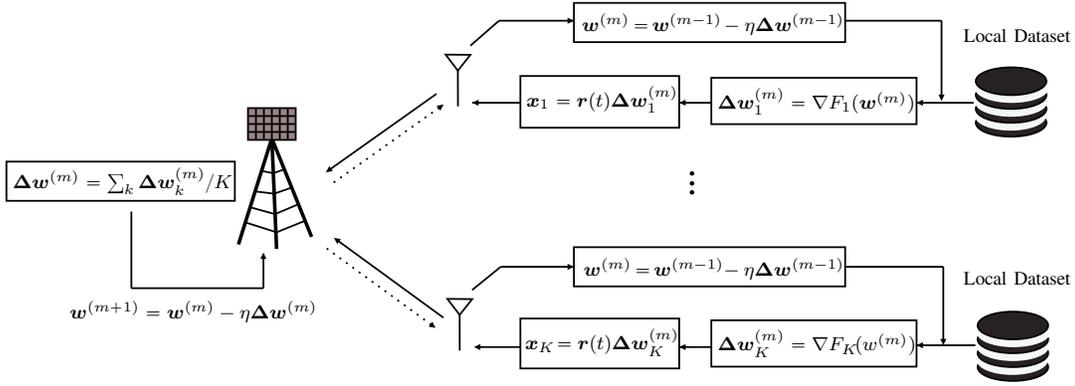

%% file: Figs/SamplingModel.tex
\begin{figure*}
    \centering

\scalebox{0.95}{

\tikzset{every picture/.style={line width=0.75pt}} 

\begin{tikzpicture}[x=0.75pt,y=0.75pt,yscale=-1,xscale=1]

\draw [line width=0.75]    (383,110) -- (408,110) ;
\draw [shift={(411,110)}, rotate = 180] [fill={rgb, 255:red, 0; green, 0; blue, 0 }  ][line width=0.08]  [draw opacity=0] (5.36,-2.57) -- (0,0) -- (5.36,2.57) -- cycle    ;
\draw [line width=0.75]    (71,91) -- (96,91) ;
\draw [shift={(99,91)}, rotate = 180] [fill={rgb, 255:red, 0; green, 0; blue, 0 }  ][line width=0.08]  [draw opacity=0] (5.36,-2.57) -- (0,0) -- (5.36,2.57) -- cycle    ;
\draw   (307,109) .. controls (307,104.03) and (311.25,100) .. (316.5,100) .. controls (321.75,100) and (326,104.03) .. (326,109) .. controls (326,113.97) and (321.75,118) .. (316.5,118) .. controls (311.25,118) and (307,113.97) .. (307,109) -- cycle ; \draw   (309.78,102.64) -- (323.22,115.36) ; \draw   (323.22,102.64) -- (309.78,115.36) ;
\draw [line width=0.75]    (316,75) -- (316.44,97) ;
\draw [shift={(316.5,100)}, rotate = 268.85] [fill={rgb, 255:red, 0; green, 0; blue, 0 }  ][line width=0.08]  [draw opacity=0] (5.36,-2.57) -- (0,0) -- (5.36,2.57) -- cycle    ;
\draw   (348.96,88.97) -- (342.11,80) -- (356,80.12) -- cycle ;
\draw    (348.96,88.97) -- (349,109) ;

\draw    (326,109) -- (349,109) ;
\draw   (382.96,89.97) -- (376.11,81) -- (390,81.12) -- cycle ;
\draw    (382.96,89.97) -- (383,110) ;

\draw [line width=0.75]    (538,109) -- (558,109) ;
\draw    (557,109) -- (588,96) ;
\draw    (585,109) -- (624,109) ;
\draw    (572.01,89.98) .. controls (592.06,100.07) and (593.12,108.62) .. (598.25,121.19) ;
\draw [shift={(599,123)}, rotate = 246.8] [color={rgb, 255:red, 0; green, 0; blue, 0 }  ][line width=0.75]    (10.93,-3.29) .. controls (6.95,-1.4) and (3.31,-0.3) .. (0,0) .. controls (3.31,0.3) and (6.95,1.4) .. (10.93,3.29)   ;
\draw [shift={(570,89)}, rotate = 25.56] [color={rgb, 255:red, 0; green, 0; blue, 0 }  ][line width=0.75]    (10.93,-3.29) .. controls (6.95,-1.4) and (3.31,-0.3) .. (0,0) .. controls (3.31,0.3) and (6.95,1.4) .. (10.93,3.29)   ;
\draw [line width=0.75]    (71,103) -- (96,103) ;
\draw [shift={(99,103)}, rotate = 180] [fill={rgb, 255:red, 0; green, 0; blue, 0 }  ][line width=0.08]  [draw opacity=0] (5.36,-2.57) -- (0,0) -- (5.36,2.57) -- cycle    ;
\draw [line width=0.75]    (71,117) -- (96,117) ;
\draw [shift={(99,117)}, rotate = 180] [fill={rgb, 255:red, 0; green, 0; blue, 0 }  ][line width=0.08]  [draw opacity=0] (5.36,-2.57) -- (0,0) -- (5.36,2.57) -- cycle    ;
\draw [line width=0.75]    (430,110) -- (455,110) ;
\draw [shift={(458,110)}, rotate = 180] [fill={rgb, 255:red, 0; green, 0; blue, 0 }  ][line width=0.08]  [draw opacity=0] (5.36,-2.57) -- (0,0) -- (5.36,2.57) -- cycle    ;
\draw [line width=0.75]    (71,132) -- (96,132) ;
\draw [shift={(99,132)}, rotate = 180] [fill={rgb, 255:red, 0; green, 0; blue, 0 }  ][line width=0.08]  [draw opacity=0] (5.36,-2.57) -- (0,0) -- (5.36,2.57) -- cycle    ;
\draw [line width=0.75]    (178,91) -- (200,91) ;
\draw [shift={(203,91)}, rotate = 180] [fill={rgb, 255:red, 0; green, 0; blue, 0 }  ][line width=0.08]  [draw opacity=0] (5.36,-2.57) -- (0,0) -- (5.36,2.57) -- cycle    ;
\draw [line width=0.75]    (178,103) -- (200,103) ;
\draw [shift={(203,103)}, rotate = 180] [fill={rgb, 255:red, 0; green, 0; blue, 0 }  ][line width=0.08]  [draw opacity=0] (5.36,-2.57) -- (0,0) -- (5.36,2.57) -- cycle    ;
\draw [line width=0.75]    (178,117) -- (200,117) ;
\draw [shift={(203,117)}, rotate = 180] [fill={rgb, 255:red, 0; green, 0; blue, 0 }  ][line width=0.08]  [draw opacity=0] (5.36,-2.57) -- (0,0) -- (5.36,2.57) -- cycle    ;
\draw [line width=0.75]    (178,132) -- (200,132) ;
\draw [shift={(203,132)}, rotate = 180] [fill={rgb, 255:red, 0; green, 0; blue, 0 }  ][line width=0.08]  [draw opacity=0] (5.36,-2.57) -- (0,0) -- (5.36,2.57) -- cycle    ;
\draw [line width=0.75]    (282,109) -- (304,109) ;
\draw [shift={(307,109)}, rotate = 180] [fill={rgb, 255:red, 0; green, 0; blue, 0 }  ][line width=0.08]  [draw opacity=0] (5.36,-2.57) -- (0,0) -- (5.36,2.57) -- cycle    ;
\draw   (411,110) .. controls (411,105.03) and (415.25,101) .. (420.5,101) .. controls (425.75,101) and (430,105.03) .. (430,110) .. controls (430,114.97) and (425.75,119) .. (420.5,119) .. controls (415.25,119) and (411,114.97) .. (411,110) -- cycle ; \draw   (413.78,103.64) -- (427.22,116.36) ; \draw   (427.22,103.64) -- (413.78,116.36) ;
\draw [line width=0.75]    (420,76) -- (420.44,98) ;
\draw [shift={(420.5,101)}, rotate = 268.85] [fill={rgb, 255:red, 0; green, 0; blue, 0 }  ][line width=0.08]  [draw opacity=0] (5.36,-2.57) -- (0,0) -- (5.36,2.57) -- cycle    ;
\draw  [color={rgb, 255:red, 21; green, 20; blue, 20 }  ,draw opacity=1 ][fill={rgb, 1:red, 0.54; green, 0.81; blue, 0.94 }  ,fill opacity=0.5 ] (458,88.59) .. controls (458,81.77) and (463.53,76.24) .. (470.35,76.24) -- (524.65,76.24) .. controls (531.47,76.24) and (537,81.77) .. (537,88.59) -- (537,125.65) .. controls (537,132.47) and (531.47,138) .. (524.65,138) -- (470.35,138) .. controls (463.53,138) and (458,132.47) .. (458,125.65) -- cycle ;
\draw  [color={rgb, 255:red, 21; green, 20; blue, 20 }  ,draw opacity=1 ][fill={rgb, 1:red, 0.54; green, 0.81; blue, 0.94 }  ,fill opacity=0.5 ] (203,91) .. controls (203,84.18) and (208.53,78.65) .. (215.35,78.65) -- (269.65,78.65) .. controls (276.47,78.65) and (282,84.18) .. (282,91) -- (282,128.06) .. controls (282,134.88) and (276.47,140.41) .. (269.65,140.41) -- (215.35,140.41) .. controls (208.53,140.41) and (203,134.88) .. (203,128.06) -- cycle ;
\draw  [color={rgb, 255:red, 21; green, 20; blue, 20 }  ,draw opacity=1 ][fill={rgb, 1:red, 0.54; green, 0.81; blue, 0.94}  ,fill opacity=0.5 ] (99,91) .. controls (99,84.18) and (104.53,78.65) .. (111.35,78.65) -- (165.65,78.65) .. controls (172.47,78.65) and (178,84.18) .. (178,91) -- (178,128.06) .. controls (178,134.88) and (172.47,140.41) .. (165.65,140.41) -- (111.35,140.41) .. controls (104.53,140.41) and (99,134.88) .. (99,128.06) -- cycle ;

\draw (33,80) node [anchor=north west][inner sep=0.75pt]  [font=\footnotesize]  {$\Delta w_{k,1}$};
\draw (627,66.4) node [anchor=north west][inner sep=0.75pt]    {$\bm{Y}_{1}$};
\draw (575,71.4) node [anchor=north west][inner sep=0.75pt]    {$1/B$};
\draw (98,87) node [anchor=north west][inner sep=0.75pt]   [align=center] {{\ Quantization}\\{\ and}\\{\ Encoding}};
\draw (240,100) node  [align=left] {{ Modulation}};
\draw (240,120) node     {$\bm{r}( t)$};
\draw (297,54) node [anchor=north west][inner sep=0.75pt]   [align=left] {{\footnotesize Carrier}};
\draw (30,126.4) node [anchor=north west][inner sep=0.75pt]  [font=\footnotesize]  {$\Delta w_{k,N}$};
\draw (401,55) node [anchor=north west][inner sep=0.75pt]   [align=left] {{\footnotesize Carrier}};
\draw (497,100) node  [align=left] {{Matched-filter}};
\draw (475,110) node [anchor=north west][inner sep=0.75pt]    {$\bm{r}^{*}( -t)$};
\draw (628,120.4) node [anchor=north west][inner sep=0.75pt]    {$\bm{Y}_{N}$};
\draw (45,95) node [anchor=north west][inner sep=0.75pt]  [font=\large]  {$\vdots $};
\draw (627,90.4) node [anchor=north west][inner sep=0.75pt]  [font=\large]  {$\vdots $};

\end{tikzpicture}

}

    \caption{The block diagram of the communication system in the over-the-air computation problem. The left side shows the architecture of the transmitter of $k$-th node, and right side architecture of the transceiver is depicted}
    \label{fig:SamplingScheme}

\end{figure*}
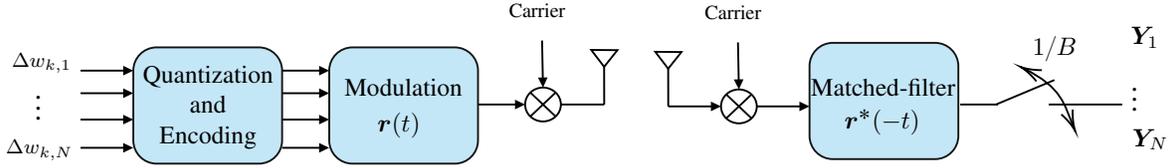

%% file: Figs/Fig_KvsSNR.tex
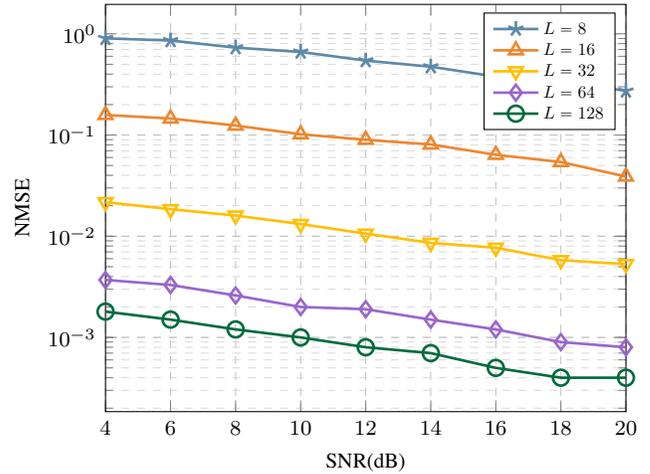
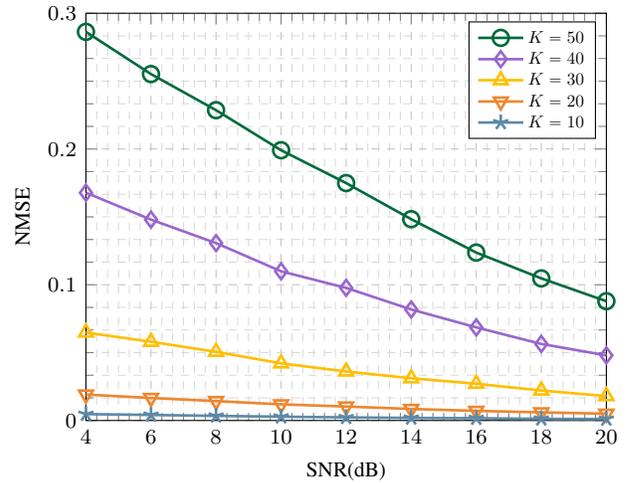
\begin{figure}[!t]
\subfigure[$K=5$ users]{ 
    \begin{tikzpicture} 
    \begin{axis}[
       xlabel={SNR(dB)},
        ylabel={NMSE},
        label style={font=\footnotesize},
        width=8.5cm,
        height=7cm,
        xmin=4, xmax=20,
        ymode = log,
        legend style={nodes={scale=0.65, transform shape}, at={(0.98,0.98)}}, 
        legend cell align={left},
        ticklabel style = {font=\footnotesize},
        ymajorgrids=true,
        xmajorgrids=true,
        grid style=dashed,
        grid=both,
        grid style={line width=.1pt, draw=gray!30},
        major grid style={line width=.2pt,draw=gray!50},
    ]
    \addplot[
        color=airforceblue,
        mark=star,
        line width=1pt,
        mark size=3pt,
        ]
    table[x=X1,y=Y5]
    {Data/NMSEvsL.dat};
    \addplot[
        color=cadmiumorange,
        mark=triangle,
        line width=1pt,
        mark size=3pt,
        ]
    table[x=X1,y=Y4]
    {Data/NMSEvsL.dat};
    \addplot[
        color=amber,
        mark=triangle,
         mark options = {rotate = 180},
        line width=1pt,
        mark size=3pt,
        ]
    table[x=X1,y=Y3]
    {Data/NMSEvsL.dat};
    \addplot[
        color=amethyst,
        mark=diamond,
        line width=1pt,
        mark size=3pt,
        ]
    ttable[x=X1,y=Y2]
    {Data/NMSEvsL.dat};
        \addplot[
        color=cadmiumgreen,
        mark=o,
        line width=1pt,
        mark size=3pt,
        ]
    table[x=X1,y=Y1]
    {Data/NMSEvsL.dat};
    \legend{$L=8$,$L=16$,$L=32$,$L=64$,$L=128$};
    \end{axis}
\end{tikzpicture}
}

\subfigure[$L=128$ samples]{ 
    \begin{tikzpicture} 
    \begin{axis}[
       xlabel={SNR(dB)},
        ylabel={NMSE},
        label style={font=\footnotesize},
        legend cell align={left},
        width=8.5cm,
        height=7cm,
        xmin=4, xmax=20,
        ymin=0, ymax=0.3,
        legend style={nodes={scale=0.65, transform shape}, at={(0.98,0.98)}}, 
        ticklabel style = {font=\footnotesize},
        ymajorgrids=true,
        xmajorgrids=true,
        grid style=dashed,
        grid=both,
        grid style={line width=.1pt, draw=gray!30},
        major grid style={line width=.2pt,draw=gray!50},
        minor tick num=5,
    ]
    \addplot[
        color=cadmiumgreen,
        mark=o,
        line width=1pt,
        mark size=3pt,
        ]
    table[x=X1,y=Y1]
    {Data/NMSEvsK.dat};
    \addplot[
        color=amethyst,
        mark=diamond,
        line width=1pt,
        mark size=3pt,
        ]
    table[x=X1,y=Y2]
    {Data/NMSEvsK.dat};
    \addplot[
        color=amber,
        mark=triangle,
        line width=1pt,
        mark size=3pt,
        ]
    table[x=X1,y=Y3]
    {Data/NMSEvsK.dat};
    \addplot[
        color=cadmiumorange,
        mark=triangle,
         mark options = {rotate = 180},
        line width=1pt,
        mark size=3pt,
        ]
    ttable[x=X1,y=Y4]
    {Data/NMSEvsK.dat};
    \addplot[
        color=airforceblue,
        mark=star,
        line width=1pt,
        mark size=3pt,
        ]
    table[x=X1,y=Y5]
    {Data/NMSEvsK.dat};
    \legend{$K=50$,$K=40$,$K=30$, $K=20$,$K=10$};
    \end{axis}
\end{tikzpicture}
}

    \caption{NMSE as a function of the SNR (dB) for (a) different number of Fourier samples $L$, and (b) different number of users $K$ by averaging over $100$ Monte Carlo trials. The results show the effect of the number of users on the performance of the proposed algorithm for estimating the parameters of the global inference model using OAC with unknown communication delays.}
    \label{fig:samples-Users_NMSE}
\end{figure}

%% file: Figs/Fig_MNIST.tex
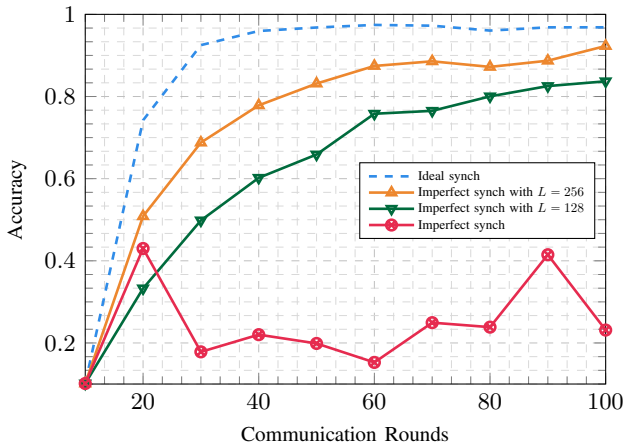
\begin{figure}[!t]
    \begin{tikzpicture} 
    \begin{axis}[
        xlabel={Communication Rounds},
        ylabel={Accuracy},
        label style={font=\footnotesize},
        legend cell align={left},
        tick label style={font=\small} , 
        width=8.5cm,
        height=6.5cm,
        xmin=10, xmax=100,
        ymin=0.1, ymax=1,
       legend style={nodes={scale=0.5, transform shape}, at={(0.98,0.6)}}, 
        ymajorgrids=true,
        xmajorgrids=true,
        grid style=dashed,
        grid=both,
        grid style={line width=.1pt, draw=gray!30},
        major grid style={line width=.2pt,draw=gray!50},
         minor tick num=5,
    ]
    \addplot[
        color=bleudefrance,
        line width=1pt,
        dashed,
        mark size=2pt,
        ]
    table[x=X1,y=Y1]
    {Data/MNIST.dat};
    \addplot[
        color=cadmiumorange,
        mark=triangle,
        line width=1pt,
        mark size=2pt,
        ]
    table[x=X1,y=Y2]
    {Data/MNIST.dat};
    \addplot[
        color=cadmiumgreen,
        mark=triangle,
        mark options = {rotate = 180},
        line width=1pt,
        mark size=2pt,
        ]
    table[x=X1,y=Y3]
    {Data/MNIST.dat};
    \addplot[
        color=amaranth,
        mark=otimes,
        line width=1pt,
        mark size=2pt,
        ]
    table[x=X1,y=Y4]
    {Data/MNIST.dat};
    \legend{Ideal synch, Imperfect synch with $L=256$, Imperfect synch with $L=128$, Imperfect synch};
    \end{axis}
\end{tikzpicture}

  \caption{Accuracy of the MNIST task as a function of the communication rounds for $K=10$ users and SNR $5$ dB. The results show the performance of our proposed recovery method for imperfect and perfect synchronization. }
  \label{fig:MNIST}
   
\end{figure}